\documentclass[journal]{IEEEtran}
\usepackage{amsmath}
\usepackage{algorithm}
\usepackage{algorithmic}
\usepackage{graphicx}
\usepackage{multirow}
\usepackage{tabularx}

\usepackage{soul}
\usepackage{colortbl}  
\usepackage{color,xcolor}
\usepackage{array}
\usepackage{amssymb}
\usepackage{bbding}
\usepackage{subfigure}
\usepackage{diagbox}
\usepackage{rotating}
\usepackage{bm}
\usepackage{booktabs}
\usepackage{cite}
\usepackage{CJK}
\usepackage{overpic}
\usepackage[colorlinks,
            linkcolor=red,
            anchorcolor=blue,
            citecolor=black
            ]{hyperref}

\newcommand{\addFig}[1]{}
\newcommand{\addFigs}[1]{}

\usepackage{subfigure}
\newcommand{\etal}{\textit{et~al}.~}
\newcommand{\ie}{\textit{i}.\textit{e}.,~}

\definecolor{lightgreen}{HTML}{9aff99}
\definecolor{lightpink}{HTML}{FFE2E1}
\definecolor{lightblue}{HTML}{c9e5ff}
\definecolor{lightpurple}{HTML}{abacf4}
\definecolor{lightorange}{HTML}{ffcc67}
\definecolor{lightgray}{HTML}{c0c0c0}
\definecolor{lightred}{HTML}{FF8181}

\soulregister{\cite}7 
\soulregister{\citep}7 
\soulregister{\citet}7 
\soulregister{\ref}7 
\soulregister{\pageref}7 

\usepackage{balance}
\usepackage{cleveref}
\crefformat{section}{\S#2#1#3} 
\crefformat{subsection}{\S#2#1#3}
\crefformat{subsubsection}{\S#2#1#3}

\usepackage{nomencl}

\usepackage{algorithm}
\usepackage{algorithmic}

\ifCLASSINFOpdf
\else
\fi

\begin{document}
%
\title{No-Service Rail Surface Defect Segmentation via Normalized Attention and Dual-scale Interaction}
%
%
%

\author{Gongyang~Li,
Chengjun Han,
and Zhi~Liu,~\IEEEmembership{Senior Member,~IEEE}
        
\thanks{Authors are all are with Key Laboratory of Specialty Fiber Optics and Optical Access Networks, Joint International Research Laboratory of Specialty Fiber Optics and Advanced Communication, Shanghai Institute for Advanced Communication and Data Science, Shanghai University, Shanghai 200444, China, and School of Communication and Information Engineering, Shanghai University, Shanghai 200444, China. Gongyang Li and Zhi Liu are also with Wenzhou Institute of Shanghai University, Wenzhou 325000, China (email: ligongyang@shu.edu.cn; monxxcn@gmail.com; liuzhisjtu@163.com).
\textit{(Gongyang Li and Chengjun Han contributed equally to this work.)}}
\thanks{\textit{Corresponding author: Zhi Liu.}}
}

\markboth{IEEE TRANSACTIONS ON INSTRUMENTATION AND MEASUREMENT}%
{Shell \MakeLowercase{\textit{et al.}}: Bare Demo of IEEEtran.cls for IEEE Journals}

\maketitle

\begin{abstract}
No-service rail surface defect (NRSD) segmentation is an essential way for perceiving the quality of no-service rails.
However, due to the complex and diverse outlines and low-contrast textures of no-service rails, existing natural image segmentation methods cannot achieve promising performance in NRSD images, especially in some unique and challenging NRSD scenes, such as low illumination, chaotic background, multiple/tiny defects, and inconsistent defect. 
To this end, in this paper, we propose a novel segmentation network for NRSDs based on Normalized Attention and Dual-scale Interaction, named \emph{NaDiNet}.
Specifically, NaDiNet follows the enhancement-interaction paradigm.
The \emph{Normalized Channel-wise Self-Attention Module} (NAM) and the \emph{Dual-scale Interaction Block} (DIB) are two key components of NaDiNet.
NAM is a specific extension of the channel-wise self-attention mechanism (CAM) to enhance features extracted from low-contrast NRSD images.
The softmax layer in CAM will produce very small correlation coefficients which are not conducive to low-contrast feature enhancement.
Instead, in NAM, we directly calculate the normalized correlation coefficient between channels to enlarge the feature differentiation.
DIB is specifically designed for the feature interaction of the enhanced features.
It has two interaction branches with dual scales, one for fine-grained clues and the other for coarse-grained clues.
With both branches working together, DIB can perceive defect regions of different granularities.
With these modules working together, our NaDiNet can generate accurate segmentation map.
Extensive experiments on the public NRSD-MN dataset with man-made and natural NRSDs demonstrate that our proposed NaDiNet with various backbones (\ie VGG, ResNet, and DenseNet) consistently outperforms 10 state-of-the-art methods.
The code and results of our method are available at https://github.com/monxxcn/NaDiNet.
\end{abstract}


\IEEEpeerreviewmaketitle

\section{Introduction}
\IEEEPARstart{S}{egmention} is a foundational and critical task in the computer vision community, aiming at assigning a category label to each pixel~\cite{2015FCN,2017SegNet,DeeplabV3,2023LASNet}.
As a pixel-wise classification task, segmentation can accurately locate the target objects in a scene.
Therefore, segmentation systems are very popular in industrial society, especially in steel companies, to determine the exact location of defects~\cite{2014Review,2020Survey,2023Review,2020MRFGMM}.
With the help of automated and precise defect segmentation systems, the industrial company can rapidly locate and repair defects to ensure the quality of its products.
Nowadays, with the popularization of high-speed rail, the reliability of railway tracks is very important.
Thus, in this paper, we focus on the No-Service Rail Surface Defect (NRSD) segmentation~\cite{2021MCNet,2022PCM,2022CLANet,2022DRERNet,2023FHENet} to ensure the reliability of the no-service rail and improve production efficiency.

As we all know, NRSD images are taken in the production workshop, usually with low lighting and chaotic environments.
In addition, the defect region also has unique characteristics such as various shapes and scales, high similarity to the non-defect region, complex and extremely irregular contours, and significant differences between different defects.
The above factors result in significant differences between natural images and NRSD images.
For example, the former has obvious textures, rich colors and clear boundaries, while the latter only has low-contrast textures, a single color and illegible boundaries.
Obviously, the advanced natural images segmentation methods~\cite{2015FCN,2015Unet,2017SegNet,DeeplabV3} are not suitable for NRSD images.

Currently, there are relatively few specialized NRSD segmentation methods available.
According to the input data, these NRSD segmentation methods can be divided into two categories, including the single image-based method~\cite{2021MCNet,2022PCM} and the RGB-D-based method~\cite{2022CLANet,2022DRERNet,2023FHENet}.
The former follows a traditional input mode, which is the category we focus on in this paper, while the latter follows a rising input mode inspired by multimodal data processing methods~\cite{20ICNet,21HAINet,20CMWNet,20BBS}.
Among these methods, MCnet~\cite{2021MCNet} is a pioneer in Convolutional Neural Networks (CNNs)~\cite{1989CNN}-based NRSD segmentation methods.
It collects a big dataset for NRSD segmentation, named NRSD-MN, containing 4,101 NRSD images.
This dataset provides a data foundation for the development of CNN-based NRSD segmentation methods.
Besides, MCnet also provides a model benchmark.
The performance of various models in the benchmark indicates that there is significant room for performance improvement in NRSD segmentation and that there is an urgent need to develop specialized CNN-based NRSD segmentation methods.

Motivated by the above observations, in this paper, we propose a novel specialized network for NRSD segmentation, termed \emph{NaDiNet}, which explores the enhancement-interaction paradigm to adapt to the unique characteristics of NRSD images.
Our main idea is to first enhance the basic features (\ie intra-level enhancement) and then interact with cross-level features (\ie inter-level interaction).
The self-attention mechanism~\cite{2017transformer,2019DANet} is a commonly used feature enhancement manner.
However, we found that the classical Channel-wise Self-Attention Mechanism (CAM) does not work well in NRSD images.
So, we analyze the presentation form of NRSD images in detail, and extend CAM to enhance features extracted from NRSD images more effectively in NaDiNet.
Furthermore, based on the enhanced features, we focus on capturing the cross-level context to perceive defect regions from multiple perspectives, which is conducive to segmenting defects completely.

Specifically, we achieve our NaDiNet in the classic encoder-decoder architecture for image segmentation~\cite{2017SegNet}.
Our NaDiNet consists of four components, including the feature extractor, the Normalized Channel-wise Self-Attention Module (NAM), the Dual-scale Interaction Block (DIB), and the segmentation head.
Notably, NAM is tailored for NRSD images to achieve effective feature enhancement.
We argue that the softmax layer in CAM significantly reduces the value of correlation coefficients in the attention map, which is detrimental to low-contrast NRSD images.
Thus, in NAM, we remove the original softmax layer of CAM and directly normalize the dependency relationships between channels to obtain the normalized correlation coefficients.
DIB is connected after NAM, and is proposed for cross-level feature interaction.
We design two interaction branches in DIB, one for large-scale interaction and the other for small-scale interaction, and capture the common knowledge of both branches for perceiving defect regions comprehensively.
In this way, our NaDiNet can segment NRSD more accurately than all compared methods and generate satisfactory segmentation maps.

Our main contributions are in three aspects:
\begin{itemize}

\item We explore the enhancement-interaction paradigm, and propose a novel specialized solution, \emph{NaDiNet}, for NRSD segmentation.
In NaDiNet, we develop an extended normalized attention and an effective dual-scale interaction manner for enhancement and interaction, respectively, resulting in our method exhibiting excellent performance on the NRSD-MN dataset, even with different backbones.

\item We extend the vanilla channel-wise self-attention mechanism, and propose the NAM to directly model the normalized long-range dependencies between channels, \ie without using the softmax layer for relative importance measurement, to effectively adapt to the unique scenes of NRSDs for feature enhancement.

\item We propose the DIB to perform cross-level feature interaction at dual scales and extract the common knowledge at both scales, comprehensively capturing the fine-grained and coarse-grained clues of NRSDs for subsequent segmentation.

\end{itemize}

The rest of this paper is organized as follows.
In Sec.~\ref{sec:related}, we review the related work of CNN-based natural image segmentation and NRSD segmentation.
In Sec.~\ref{sec:OurMethod}, we elaborate our method in detail.
In Sec.~\ref{sec:exp}, we present experiments, ablation studies, and analysis.
In Sec.~\ref{sec:con}, we summarize the conclusion.

\section{Related Work}
\label{sec:related}
In this section, we review the representative works of CNN-based natural image segmentation and NRSD segmentation.

\subsection{CNN-based Natural Image Segmentation}
\label{sec:NSI_Seg}
With the use of CNNs in natural image segmentation, the segmentation performance has been greatly improved.
As a pioneer work in the field of segmentation based on CNNs, Long~\etal\cite{2015FCN} proposed first end-to-end CNN-based segmentation method, namely Fully Convolutional Network (FCN).
FCN included an effective skip architecture for integrating multi-level information for accurate segmentation.
As a contemporaneous work, Noh~\etal\cite{2015DeconvNet} proposed the Deconvolution Network (DeconvNet), which includes not only a convolution network, but also an ingenious deconvolution network.
The deconvolution network of DeconvNet gradually inferred the object through the unpooling and deconvolution operations.
Similarly, Badrinarayanan~\etal\cite{2017SegNet} proposed the famous SegNet, which is an encoder-decoder architecture, for image segmentation.

The above three efforts have laid the foundation for CNN-based natural image segmentation.
Subsequent segmentation methods are mainly developed from multi-scale perception and context information extraction.
For example, Chen~\etal\cite{DeepLabV1,DeepLabV2,DeeplabV3,DeepLabV3+} proposed the Deeplab series of segmentation methods, and perceived multi-scale information with atrous convolutions (\ie atrous spatial pyramid pooling (ASPP)).
Yang~\etal\cite{2018DenseASPP} introduced the densely connected structure into ASPP, proposing DenseASPP to achieve the multi-scale perception with a very large scale range.
In~\cite{2017PSP}, Zhao~\etal proposed the pyramid parsing module to perceive different sub-region representations using pooling layers.
For context information extraction, Yuan~\etal\cite{2018OCNet} adopted the self-attention mechanism to extract the object context.
Moreover, Fu~\etal\cite{2019DANet} applied the self-attention mechanism not only in the position domain, but also in the channel domain, proposing DANet to integrate local features with global contextual dependencies.
Differently, Huang~\etal\cite{2019CCNet} harvested the context information in a criss-cross manner in their proposed criss-cross attention.
Driven by Transformers~\cite{2017transformer,2021ViT}, Xie~\etal\cite{2021Segformer} and Strudel~\etal\cite{2021Segmenter} modeled and captured the contextual information more powerful, and proposed SegFormer and Segmenter, respectively.

Due to differences in the photography environment and shooting targets, the above CNN-based segmentation methods for natural images are not suitable for NRSD images we care about.
But these methods give us some inspiration.
Concretely, our solution is based on the famous encoder-decoder architecture\cite{2017SegNet}.
And we extend the traditional self-attention mechanism\cite{2017transformer,2019DANet}, which is widely used in natural image segmentation, to NAM for feature enhancement in our solution.
Our NAM is designed specifically for NRSD images and can alleviate the shortcoming of CAM's inability to effectively enhance low-contrast NRSD images.

\subsection{No-Service Rail Surface Defect Segmentation}
\label{sec:RSD_SDI}
NRSD segmentation can achieve pixel-level defect localization, which can significantly improve the quality of no-service rail.
With the development of technology, the segmentation system has also been upgraded from a single camera to a complex system composed of multiple data acquisition devices, resulting in two categories of methods for NRSD segmentation.
The first one is based on a single NRSD image.
To solve the problem of data shortage, Zhang~\etal\cite{2021MCNet} constructed a big dataset for NRSD segmentation.
They explored the multiple types of context information for NRSD segmentation with the pixel-wise contextual attention~\cite{2018PiCANet}.
Moreover, to alleviate the problem of expensive pixel-level annotations, Zhang~\etal\cite{2022PCM} also explored NRSD segmentation with image-level annotations.
The second one handles RGB-D data, and aims to achieve accurate segmentation with the help of the depth map.
Wang~\etal\cite{2022CLANet} explored NRSD segmentation for the first time, and mined the multi-modal complementarity of multiple levels in a multimodal attention block.
Wu~\etal\cite{2022DRERNet} repeatedly used multi-modal features to locate defects and refine their boundaries.
Zhou~\etal\cite{2023FHENet} handled multi-modal and multi-level feature integration via a hierarchical exploration strategy.

In addition to the above NRSD segmentation methods, we also introduce segmentation methods for other materials similar to no-service rail, such as steel~\cite{2019ADSD,2022DADL,2023MFIP},
steel sheet~\cite{2018DLR,2021JCS},
and strip steel~\cite{2020EDRNet,2021EMINet,2021DACNet,2022CSEP,2022TSERNet,2023FewShotSeg}.
For steel, Qian~\etal\cite{2019ADSD} adopted the model ensemble strategy to achieve good performance in both time and accuracy.
Pan~\etal\cite{2022DADL} integrated the position and channel self-attention mechanisms into Deeplabv3+, proposing DAN-Deeplabv3+ for steel defect segmentation.
Zhao~\etal\cite{2023MFIP} captured the intrinsic relationship of features at different levels to adapt to the complexity of steel defects.
For steel sheet, Zhou~\etal\cite{2018DLR} regarded defect regions as the salient part of image, and proposed a saliency detection method based on double low-rank and sparse decomposition to segment defects.
In~\cite{2021JCS}, they implemented the approach of classification first and then segmentation.
For strip steel, similar to~\cite{2018DLR}, many researchers also regarded the surface defects of strip steel as salient regions, and proposed a series of saliency detection methods for strip steel, which explore the prediction-refinement strategy~\cite{2020EDRNet,2022TSERNet}, the edge information~\cite{2021EMINet,2022CSEP,2022TSERNet}, and the multi-resolution inputs~\cite{2021DACNet}.
Moreover, Feng~\etal\cite{2023FewShotSeg} achieved few-shot strip steel surface defect segmentation to improve the performance of defects with insufficient samples and unseen defects.

In general, the specialized methods for NRSD segmentation are quite scarce, especially for the single image-based NRSD segmentation.
For MCnet, the only method of this category, it employs the existing pixel-wise contextual attention module to handle the unique NRSD images, which is a suboptimal solution.
In our NaDiNet, based on the characteristics of NRSD images, we propose the NAM to enlarge the feature differentiation of the low-contrast features for effective feature enhancement.
Our NAM is superior to the contextual attention module (proposed for natural images) used in MCnet.
In addition to the feature enhancement module, we propose a tailored DIB instead of the simple feature summation used in MCnet to achieve cross-level feature interaction.
Our DIB can capture multi-scale information of defect regions, which is beneficial for handling scenes with multiple defects and defects of various sizes in NRSD images.
The above efforts result in our NaDiNet substantially ahead of state-of-the-art methods, including MCnet.

\begin{figure}
	\centering
	\begin{overpic}[width=1\columnwidth]{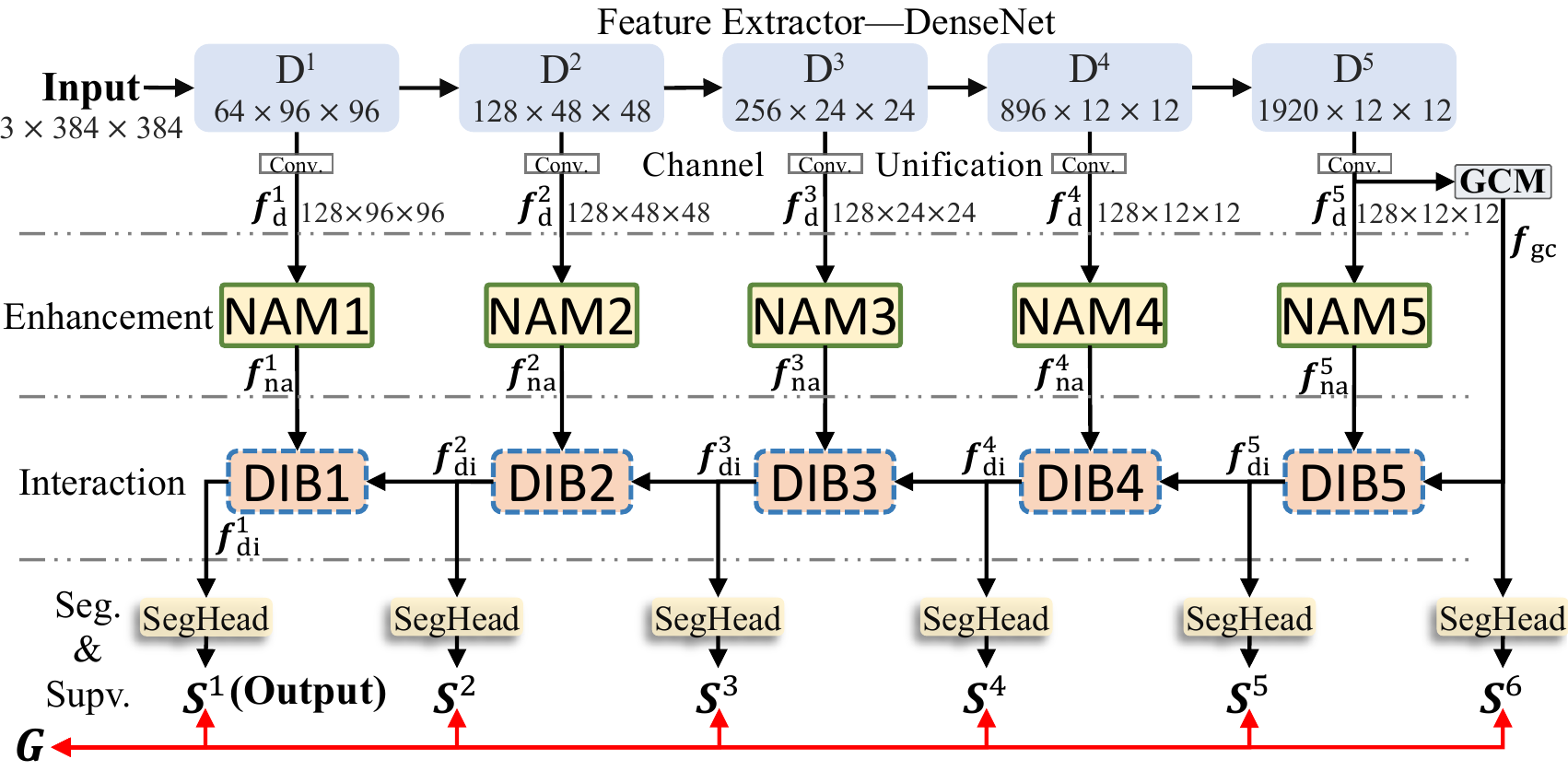}
    \end{overpic}
	\caption{Pipeline of the proposed NaDiNet for NRSD segmentation.
	NaDiNet follows the enhancement-interaction paradigm, and consists of the feature extractor, the Normalized Channel-wise Self-Attention Module (NAM), the Dual-scale Interaction Block (DIB), and the segmentation head.
	The feature extractor is comprised of the DenseNet-201~\cite{DenseNet} and a Global Contextual Module (GCM)~\cite{20BBS}, producing five-level basic features and the global context features, respectively.
	 Then, NAM is connected after the feature extractor, and enhances the basic features with normalized attention.
	 DIB is connected after NAM, and performs cross-level feature interaction at dual scales.
	 Finally, the segmentation head is connected after the DIB/GCM, and generates the segmentation map.
	 Notably, there are six segmentation heads in our NaDiNet.
	 The segmentation head after DIB1 generates the output segmentation map of our NaDiNet, while the other segmentation heads are used for deep supervision during the network training phase.
    }
    \label{fig:Framework}
\end{figure}

\section{Proposed Method}
\label{sec:OurMethod}
In this section, we present our NaDiNet in detail.
In Sec.~\ref{sec:Overview}, we introduce the network overview of our NaDiNet.
In Sec.~\ref{sec:NAM}, we elaborate the proposed NAM.
In Sec.~\ref{sec:DIB}, we elaborate the proposed DIB.
In Sec.~\ref{sec:Loss}, we formulate the loss function.

\subsection{Network Overview}
\label{sec:Overview}
As illustrated in Fig.~\ref{fig:Framework}, the proposed NaDiNet follows the enhancement-interaction paradigm, and includes a feature extractor, five NAMs, five DIB, and six segmentation heads.
For feature extractor, we take the DenseNet-201~\cite{DenseNet} as an example here, which can be replaced by other backbones, such as VGG~\cite{2014VGG16} and ResNet~\cite{2016ResNet}.
We arrange a convolutional layer with ReLU activation function after each block (denoted as D$^{i}$ ($i=1,2,3,4,5$)) of DenseNet-201 for feature channel unification, extracting five-level basic features $\boldsymbol{{f}}^{i}_\mathrm{d} \in \mathbb{R}^{c_i\!\times\!h_i\!\times\!{w}_i}$ where $c_i$ is 128.
Since the size of input NRSD image is $384\!\times\!384$, $h_i$ and $w_i$ belong to $\{96,48,24,12,12\}$.
And the feature extractor also includes a Global Contextual Module (GCM)~\cite{20BBS}, which contains four parallel convolutional layers with progressive dilation rates, to extract global context features $\boldsymbol{{f}}_\mathrm{gc} \in \mathbb{R}^{c_5\!\times\!h_5\!\times\!{w}_5}$.
Then, in the NAM, we perform feature enhancement on $\boldsymbol{{f}}^{i}_\mathrm{d}$ by the normalized attention, which is tailored for the low-contrast textures of NRSD images, generating $\boldsymbol{{f}}^{i}_\mathrm{na} \in \mathbb{R}^{c_i\!\times\!h_i\!\times\!{w}_i}$.
Next, in DIB, we carry out the dual-scale cross-level feature interaction, that is, we integrate cross-level features on a large scale and a small scale to perceive defect regions of different granularities, generating $\boldsymbol{{f}}^{i}_\mathrm{di} \in \mathbb{R}^{c_i\!\times\!h_i\!\times\!{w}_i}$.
Except for DIB5 which deals with $\boldsymbol{{f}}^{5}_\mathrm{na}$ and $\boldsymbol{{f}}_\mathrm{gc}$, the rest of DIBs are used to interact $\boldsymbol{{f}}^{i}_\mathrm{na}$ and $\boldsymbol{{f}}^{i+1}_\mathrm{di}$.
Finally, the segmentation head (SegHead) consisting of a convolutional layer without the activation function and an upsampling operation is used to generate the segmentation map and restore its size to $384\!\times\!384$.
In our NaDiNet, the segmentation map generated by the SegHead after DIB1, \ie $\boldsymbol{S}^{1} \in \mathbb{R}^{1\!\times\!384\!\times\!384}$, is the output segmentation map of our NaDiNet.

\begin{figure}
\centering
\footnotesize
  \begin{overpic}[width=1\columnwidth]{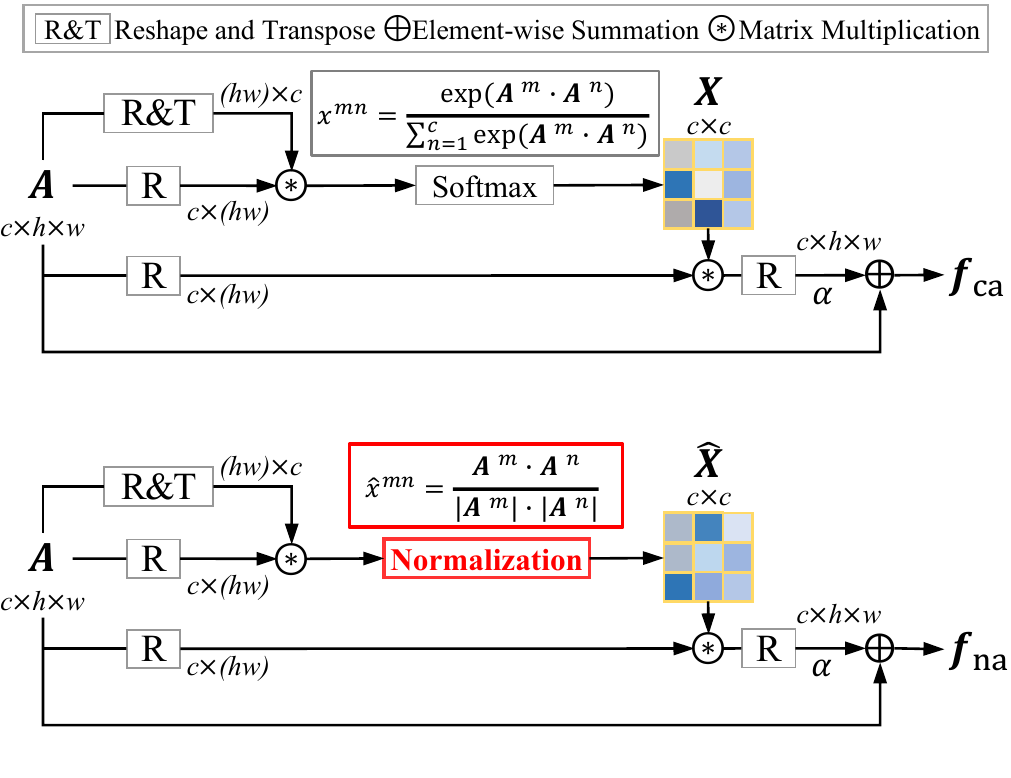}
    \put(11.75,37.80){\textbf{(a) Vanilla channel-wise self-attention mechanism (CAM)}}
    \put(8.02,1.10){\textbf{(b) Our normalized channel-wise self-attention module (NAM)}}
  \end{overpic}
\caption{
Illustration of vanilla CAM and our proposed NAM.
}
\label{NAM_structure}
\end{figure}

\begin{figure}
\centering
\footnotesize
  \begin{overpic}[width=1\columnwidth]{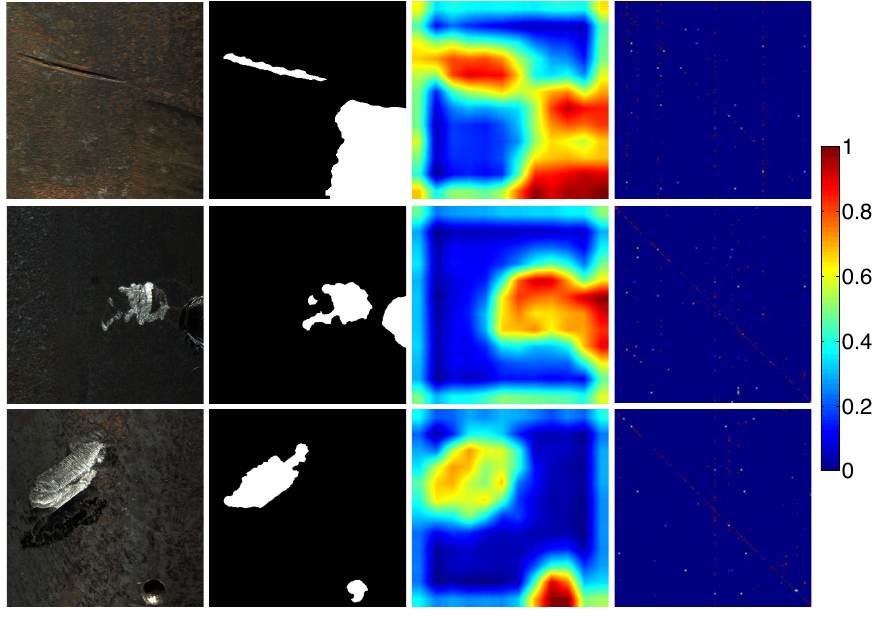}
    \put(3.3,-0.8){NRSD image}
    \put(32.8,-0.8){GT}
    \put(55.9,-1.1){$\boldsymbol{f}^{4}_\mathrm{ca}$}
    \put(79.5,-1.1){$\boldsymbol{X}^{4}$}
  \end{overpic}
\caption{Feature visualization of the enhanced feature $\boldsymbol{f}^{4}_\mathrm{ca}$ and the channel attention map $\boldsymbol{X}^{4}$ in vanilla CAM4. Zoom-in for viewing details.}
\label{CAM_Visualization}
\end{figure}
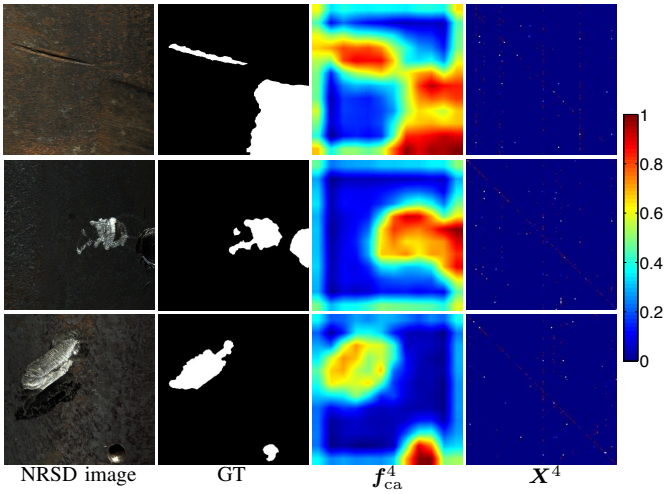

\subsection{Normalized Channel-wise Self-Attention Module}
\label{sec:NAM} 
In this section, we first elaborate the vanilla Channel-wise Self-Attention Mechanism (CAM) of DANet~\cite{2019DANet}, which is the basis of our NAM.
We illustrate the detailed structure of CAM in Fig.~\ref{NAM_structure} (a).
Here, for simplicity, we take $\boldsymbol{A} \in \mathbb{R}^{c\!\times\!h\!\times\!w}$ as the input feature of CAM. 
Specifically, we reshape $\boldsymbol{A}$ to $\mathbb{R}^{c\!\times\!(hw)}$, and then perform the matrix multiplication between the reshaped $\boldsymbol{A}$ and the transpose of the reshaped $\boldsymbol{A}$.
Next, we adopt a softmax layer to produce the channel attention map $\boldsymbol{X} \in \mathbb{R}^{c\!\times\!c}$ of CAM, which can be formulated as:
\begin{equation}
   \begin{aligned}
	x^{mn} = \frac{\textrm{exp}(\textit{\textbf{A}}^{m} \cdot \textit{\textbf{A}}^{n})}{\sum_{n=1}^{c} \textrm{exp}(\textit{\textbf{A}}^{m} \cdot \textit{\textbf{A}}^{n}) },
    \label{eq:cam}
    \end{aligned}
\end{equation}
where $x^{mn}$ measures the correlation between the \textit{n-th} channel and the \textit{m-th} channel, and $x^{mn} \in(0,1)$ and $\sum_{n=1}^{c} x^{mn} = 1$.
In addition, we transfer the correlation of $\boldsymbol{X}$ to the reshaped $\boldsymbol{A}$ by the matrix multiplication, and reshape the generated features to $\mathbb{R}^{c\!\times\!h\!\times\!w}$.
Finally, we fuse the reshaped features and the input $\boldsymbol{A}$ via an element-wise summation with a coefficient (\ie $\alpha$) to generate the enhanced feature $\boldsymbol{{f}}_\mathrm{ca} \in \mathbb{R}^{c\!\times\!h\!\times\!w}$.

CAM is designed for natural images with rich texture and color cues.
Since NRSD images are usually with low-contrast textures and the single dull tone, CAM is a suboptimal feature enhancement method for NRSD images.
For intuitive understanding, we use the CAM in our NaDiNet for feature enhancement, and visualize the output feature $\boldsymbol{{f}}_\mathrm{ca}^4$ of CAM4 in Fig.~\ref{CAM_Visualization}.
We observe that defect regions are only highlighted roughly in $\boldsymbol{{f}}_\mathrm{ca}^4$, and even the two defect regions in the second example are connected together, which is not accurate and fine enough.
We argue that the generation approach of the channel attention map makes CAM not suitable for NRSD images with low-contrast textures.
Concretely, the softmax layer used for relative importance measurement will make $\sum_{n=1}^{c} x^{mn}$ to 1, making the value of each pixel in $\boldsymbol{X}$ very small.
Such small values are unfriendly to the low-contrast NRSD images, and cannot clearly highlight the valuable channels.
The above inference is also consistent with the corresponding channel attention map $\boldsymbol{X}^4$ of CAM4 shown in the last column of Fig.~\ref{CAM_Visualization}.
For these three examples, $\boldsymbol{X}^4$ is almost all blue, which indicates that the values of $\boldsymbol{X}^4$ are close to 0, that is, $\boldsymbol{X}^4$ cannot effectively model the relationship between channels for NRSD images.
Thus, $\boldsymbol{{f}}_\mathrm{ca}^4$ is not effectively enhanced.

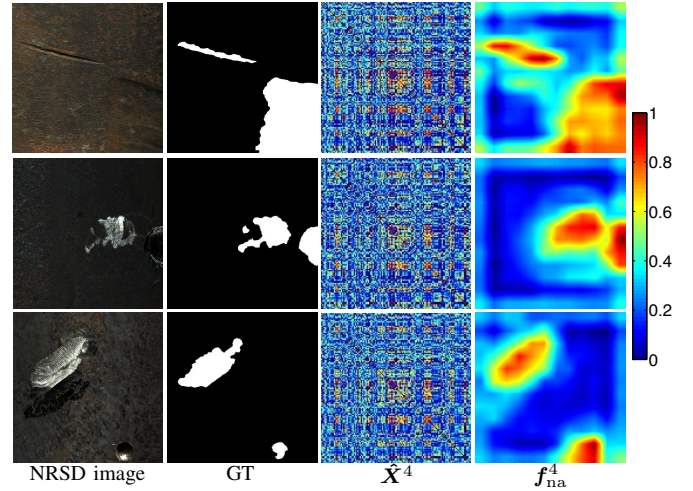
\begin{figure}
\centering
\footnotesize
  \begin{overpic}[width=1\columnwidth]{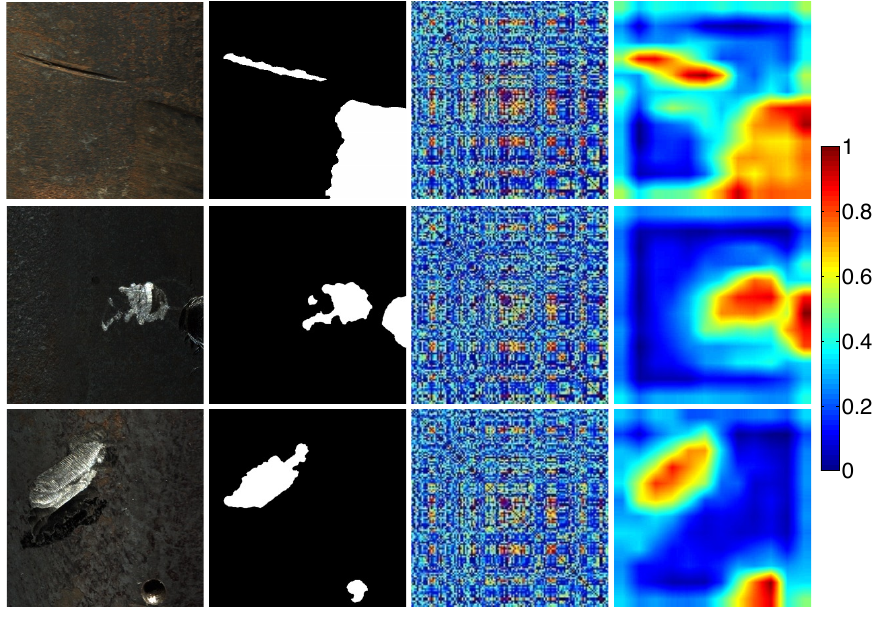}
    \put(3.3,-0.8){NRSD image}
    \put(32.8,-0.8){GT}
    \put(55.9,-1.3){$\boldsymbol{\hat{X}}^{4}$}
    \put(78.9,-1.1){$\boldsymbol{f}^{4}_\mathrm{na}$}
  \end{overpic} 
\caption{Feature visualization of the channel attention map $\boldsymbol{\hat{X}}^{4}$ and the enhanced feature $\boldsymbol{f}^{4}_\mathrm{na}$ in the proposed NAM4. Zoom-in for viewing details.}
\label{NAM_Visualization}
\end{figure}

%
Accordingly, we propose NAM to directly model the normalized relationship between two channels to enlarge the feature differentiation and achieve effective enhancement for NRSD features, generating the enhanced feature $\boldsymbol{{f}}_\mathrm{na} \in \mathbb{R}^{c\!\times\!h\!\times\!w}$.
The detailed structure of NAM is illustrated in Fig.~\ref{NAM_structure} (b).
Specifically, in NAM, we propose to replace the softmax layer with the normalization operation to produce an updated channel attention map $\boldsymbol{\hat{X}} \in \mathbb{R}^{c\!\times\!c}$:
\begin{equation}
   \begin{aligned}
	\hat{x}^{mn} = \frac{\textit{\textbf{A}}^{m} \cdot \textit{\textbf{A}}^{n}} {|\textit{\textbf{A}}^{m}| \cdot |\textit{\textbf{A}}^{n}|},
    \label{eq:nam}
    \end{aligned}
\end{equation}
where $\hat{x}^{mn}$ measures normalized correlation the \textit{n-th} channel and the \textit{m-th} channel.
Notably, in our NaDiNet, the input feature $\boldsymbol{A}$ of NAM is actually $\boldsymbol{f}_\mathrm{d}$ generated by a convolutional layer with ReLU activation function, resulting in $\boldsymbol{A}$ being non-negative.
So $\hat{x}^{mn}$ belongs to $[0,1]$ without the restriction of the sum.
We present the updated channel attention map $\boldsymbol{\hat{X}}^4$ of NAM4 in Fig.~\ref{NAM_Visualization}.
Compared with $\boldsymbol{X}^4$, the values of our $\boldsymbol{\hat{X}}^4$ are significantly large.
The attention map with large values is indeed suitable for NSRD features.
It can enlarge feature channel differences using large values to handle the unique low-contrast characteristics of NSRD features, enabling effective feature enhancement.
We also show the enhanced $\boldsymbol{f}^{4}_\mathrm{na}$ generated from our NAM4 in the last column of Fig.~\ref{NAM_Visualization}.
We can clearly observe that the defect regions of all three examples are highlighted more accurately and finely than those in CAM4.

\begin{figure}
\centering
\footnotesize
  \begin{overpic}[width=0.9\columnwidth]{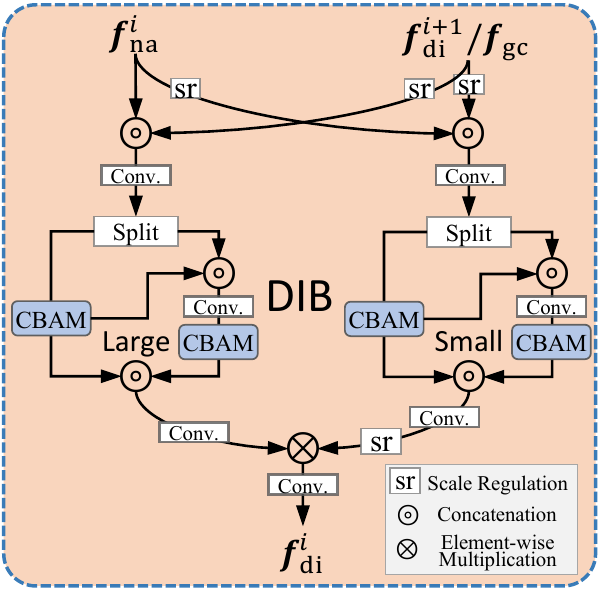}
  \end{overpic}
\caption{Illustration of the Dual-scale Interaction Block (DIB).
CBAM is the classic Convolutional Block Attention Module~\cite{cbam} with channel attention and spatial attention.}
\label{DIB_structure}
\end{figure}

\subsection{Dual-scale Interaction Block}
\label{sec:DIB}
NAM only performs the intra-level enhancement, which is not enough for NSRDs with various sizes and shapes.
We propose DIB to explore the information between different levels (\ie inter-level interaction).
Moreover, we achieve the inter-level interaction at dual scales to perceive defect regions of different granularities.
As presented in Fig.~\ref{DIB_structure}, DIB contains a large-scale interaction branch (\ie the left branch in Fig.~\ref{DIB_structure}) and a small-scale interaction branch (\ie the right branch in Fig.~\ref{DIB_structure}).
According to the input, we generally formulate the processing of DIB as $\mathrm{F} (\cdot)$:
\begin{equation}
   \begin{aligned}
    \boldsymbol{f}^{i}_{\mathrm{di}}=\left\{
	\begin{array}{lll}
	\mathrm{F} ( \boldsymbol{f}^{i}_\mathrm{na}, \boldsymbol{f}^{i+1}_\mathrm{di} ) ,     & i=1,2,3,4,\\
	\mathrm{F} ( \boldsymbol{f}^{i}_\mathrm{na}, \boldsymbol{f}_\mathrm{gc} ) ,      & i=5,\\
	\end{array}  \right. 
    \label{eq:dib}
    \end{aligned}
\end{equation}
where $\boldsymbol{f}^{i}_{\mathrm{di}}$ is the output feature of DIB.
In the following, we introduce these two interaction branch in turn.

\textit{1) Large-scale Interaction Branch.}
In the large-scale interaction branch of DIB$i$, we operate features on the size of $h_i \times w_i$.
For DIB4 and DIB5, their inputs are with the same size of $h_i \times w_i$, so we forego the scale regulation operation on $\boldsymbol{f}^{5}_\mathrm{na}$ and $\boldsymbol{f}_\mathrm{gc}$.
For the other DIBs, we adopt the upsampling operation as the scale regulation operation, resizing $\boldsymbol{f}^{i+1}_\mathrm{di}$ to the size of $h_i \times w_i$.
Then, we concatenate the inputs of this large-scale interaction branch, and use a convolutional layer to fuse them, generating $\boldsymbol{{f}}^{i}_\mathrm{b} \in \mathbb{R}^{2c_i\!\times\!h_i\!\times\!w_i}$.
Here, we perform a progressive ensemble manner to group and asymptotically modulate $\boldsymbol{{f}}^{i}_\mathrm{b}$ with the classic Convolutional Block Attention Module (CBAM)~\cite{cbam}.
Concretely, we firstly split $\boldsymbol{{f}}^{i}_\mathrm{b}$ into $\boldsymbol{{f}}^{i}_\mathrm{b1} \in \mathbb{R}^{c_i\!\times\!h_i\!\times\!w_i}$ and $\boldsymbol{{f}}^{i}_\mathrm{b2}\in \mathbb{R}^{c_i\!\times\!h_i\!\times\!w_i}$.
$\boldsymbol{{f}}^{i}_\mathrm{b1}$ is modulated through the CBAM to highlight defect regions, getting $\boldsymbol{\hat{f}}^{i}_\mathrm{b1} \in \mathbb{R}^{c_i\!\times\!h_i\!\times\!w_i}$. 
We concatenate $\boldsymbol{{f}}^{i}_\mathrm{b2}$ and $\boldsymbol{\hat{f}}^{i}_\mathrm{b1}$, and further modulate their combinations through the CBAM, producing $\boldsymbol{\hat{f}}^{i}_\mathrm{b2} \in \mathbb{R}^{c_i\!\times\!h_i\!\times\!w_i}$.
Next, we integrate $\boldsymbol{\hat{f}}^{i}_\mathrm{b1}$ and $\boldsymbol{\hat{f}}^{i}_\mathrm{b2}$ through the concatenation operation and a convolutional layer, generating the output of the large-scale interaction branch $\boldsymbol{\hat{f}}^{i}_\mathrm{b} \in \mathbb{R}^{c_i\!\times\!h_i\!\times\!w_i}$.
We formulate the above operations as follows:
\begin{equation}
   \begin{aligned}
    \left\{
	\begin{array}{lll}
	\boldsymbol{\hat{f}}^{i}_\mathrm{b1} = \mathrm{CBAM}(\boldsymbol{{f}}^{i}_\mathrm{b1}),\\
	 \boldsymbol{\hat{f}}^{i}_\mathrm{b2} = \mathrm{CBAM}(\boldsymbol{{f}}^{i}_\mathrm{b2} \circledcirc \boldsymbol{\hat{f}}^{i}_\mathrm{b1}),\\
	 \boldsymbol{\hat{f}}^{i}_\mathrm{b} =  \boldsymbol{\hat{f}}^{i}_\mathrm{b1} \circledcirc \boldsymbol{\hat{f}}^{i}_\mathrm{b2},\\
	\end{array}  \right. 
    \label{eq:big}
    \end{aligned}
\end{equation}
where $\circledcirc$ contains a concatenation operator and a convolutional layer.
In this way, we adopt two CBAMs to achieve the gradual expansion of the receptive field in a series-like manner, which can capture multi-scale information of NSRDs.

\textit{2) Small-scale Interaction Branch.}
In the small-scale interaction branch, we operate features on the size of $\frac{h_i}{2} \times \frac{w_i}{2}$.
For DIB4 and DIB5, we adopt the downsampling operation as the scale regulation operation, resizing the size of both inputs from $h_i \times w_i$ to $\frac{h_i}{2} \times \frac{w_i}{2}$.
For the other DIBs, we only apply the scale regulation operation which is the downsampling operation to $\boldsymbol{f}^{i}_\mathrm{na}$, resizing its size to $\frac{h_i}{2} \times \frac{w_i}{2}$.
And we forego the scale regulation operation on $\boldsymbol{f}^{i+1}_\mathrm{di}$.
Then, the subsequent operations of this branch are the same as those of the large-scale interaction branch, as formulated in Eq.~\ref{eq:big}.
In this way, this branch generates the output $\boldsymbol{\hat{f}}^{i}_\mathrm{s} \in \mathbb{R}^{c_i\!\times\!\frac{h_i}{2} \times \frac{w_i}{2}}$.

\textit{3) Branch Integration.}
Benefitting from the progressive ensemble manner in both interaction branches, the large one can capture the multi-scale fine-grained information, while the small one can capture the multi-scale coarse-grained information.
With the above information of different granularities, we adopt the element-wise multiplication operation to extract the common knowledge of $\boldsymbol{\hat{f}}^{i}_\mathrm{b}$ and the upsampled $\boldsymbol{\hat{f}}^{i}_\mathrm{s}$, generating the output of DIB $\boldsymbol{{f}}^{i}_\mathrm{di}$.
In this way, $\boldsymbol{{f}}^{i}_\mathrm{di}$ enables a complete characterization of defect regions, which facilitates the segmentation process in the SegHead.

\begin{table*}[t!]
  \centering
  \small
  \renewcommand{\arraystretch}{1.2}
  \renewcommand{\tabcolsep}{1.8mm}
  \caption{
    Quantitative results (\%) of 10 state-of-the-art methods with different backbones and our method with three different backbones (\ie VGG, ResNet, and DenseNet) on the NRSD-MN dataset, including the man-made NRSDs and the natural NRSDs.
    $\uparrow$ means that the bigger the better.
    The top two results are highlighted in \textbf{bold} and \underline{underline}.
    }
\label{table:QuantitativeResults}
\begin{tabular}{r|c|c|c|ccc|ccc}
\midrule[1pt]
 \multirow{2}{*}{{Methods}}
 & \multirow{2}{*}{{Backbone}}
 & \#Param & Speed
 & \multicolumn{3}{c|}{Man-made (965)}
 & \multicolumn{3}{c}{Natural (165)} \\
         &  & (M)$\downarrow$ & (\textit{fps})$\uparrow$ 
         & PA$\uparrow$ & IoU$\uparrow$ & F1$\uparrow$ & PA$\uparrow$ & IoU$\uparrow$ & F1$\uparrow$ \\
\midrule[1pt]
SegNet~\cite{2017SegNet}	& VGG & 29.44 & 29 & 75.8 & 62.3 & 90.8 & 57.6 & 50.8 & 70.5 \\
FCN~\cite{2015FCN}		& VGG & 18.64 & 31 & 79.3 & 64.8 & 92.0 & 59.9 & 53.3 & 72.5 \\
UNet~\cite{2015Unet}		& VGG & 31.04 & 78 & 79.8 & 66.4 & 93.7 & 64.0 & 57.2 & 76.1 \\
CSEPNet~\cite{2022CSEP}	& VGG & 18.78 & 12 & 81.1 & 69.3 & 95.1 & 65.1 & 59.1 & 75.2 \\ 
\textbf{Ours-V}				& VGG & 27.68 & 27 & \underline{83.9} & \textbf{71.4} & \textbf{97.5} & \underline{71.7} & \underline{63.8} & 80.0 \\
\hline
\hline
DACNet\cite{2021DACNet}	& ResNet & 98.39 & 17 & 77.0 & 66.3 & 93.7 & 61.7 & 55.4 & 72.7 \\ 
Deeplabv3~\cite{DeeplabV3}	& ResNet & 16.48 & 27 & 81.1 & 67.4 & 93.1 & 63.7 & 57.3 & 75.3 \\
TSERNet~\cite{2022TSERNet}	& ResNet & 189.64 & 21 & 80.7 & 68.7 & 93.9 & 66.2 & 59.3 & 75.7 \\
EDRNet\cite{2020EDRNet}	& ResNet & 39.31 & 17 & 81.9 & 69.0 & 95.2 & 68.3 & 60.5 & 77.8 \\
EMINet~\cite{2021EMINet}	& ResNet & 99.13 & 22 & 81.7 & 69.6 & 96.2 & 68.6 & 60.7 & 79.0 \\
\textbf{Ours-R}				& ResNet & 48.39 & 22 & 83.6 & 69.7 & 96.0 & 70.8 & 62.7 & \underline{80.9} \\
\hline
\hline
MCnet~\cite{2021MCNet} 		& DenseNet & 38.41 & 18 & 79.4 & 68.1 & 94.4 & 69.5 & 62.6 & 79.8 \\
\textbf{Ours-D}				& DenseNet & 33.11 & 13 & \textbf{84.2} & \underline{71.3} & \underline{97.2} & \textbf{72.5} & \textbf{65.2} & \textbf{82.5} \\

\toprule[1pt]

\end{tabular}

\end{table*}

\subsection{Loss Function}
\label{sec:Loss}
As shown in Fig.~\ref{fig:Framework}, there are six SegHeads attached after DIBs and GCM.
As mentioned in Sec.~\ref{sec:Overview}, the SegHead after DIB1 generates the output segmentation map of our NaDiNet, while the other SegHeads generate the side segmentation maps which are only used for deep supervision~\cite{DSNet,2022ACCoNet}.
And all segmentation maps are supervised by the ground truth binary map.
Therefore, the total loss ${L}_{\mathrm{total}}$ of our NaDiNet in the training phase can be calculated as:
\begin{equation}
   \begin{aligned}
    {L}_{\mathrm{total}} =  \sum_{i = 1}^{6}  \ell_{\mathrm{seg}}^{i} (\boldsymbol{S}^{i}, \boldsymbol{G}),
    \label{eq:Loss1}
    \end{aligned}
\end{equation}
where $\boldsymbol{S}^{i}$ is the predicted segmentation map generated from SegHead, and $\boldsymbol{G}$ is the binary ground truth map.
These two maps have the same resolution of $1\times384\times384$.
To effectively train our NaDiNet and improve the segmentation performance, $\ell_{\mathrm{seg}}$ includes not only the general pixel-level binary cross-entropy loss function $\ell_{\mathrm{bce}}$, but also the map-level intersection over union loss function $\ell_{\mathrm{iou}}$.
Thus, ${L}_{\mathrm{total}}$ is updated as:
\begin{equation}
   \begin{aligned}
    {L}_{\mathrm{total}} =  \sum_{i = 1}^{6}  \big( \ell_{\mathrm{bce}}^{i} (\boldsymbol{S}^{i}, \boldsymbol{G}) + \ell_{\mathrm{iou}}^{i} (\boldsymbol{S}^{i}, \boldsymbol{G}) \big).
    \label{eq:Loss2}
    \end{aligned}
\end{equation}
When performing segmentation reasoning after the network training is completed, we will discard the last five SegHeads and only keep the first one, whose output $\boldsymbol{S}^{1}$ is the segmentation result of our NaDiNet.

\section{Experiments}
\label{sec:exp}

\subsection{Experimental Protocol}
\label{sec:ExpProtocol}

\textit{1) Datasets.}
We conduct all experiments on the public NRSD-MN dataset~\cite{2021MCNet}.
NRSD-MN dataset contains 4,101 NRSD images and corresponding pixel-level ground truths (GTs), of which 3,936 images are man-made NRSD images and 165 images are natural NRSD images.
For the man-made NRSD images in MCnet~\cite{2021MCNet}, 2,086 images are used for training, 885 for validation, and 965 images for testing.
We only use the above training set of 2,086 images and not the validation set during network training.
After training, we test the model on the man-made NRSD testing set of 965 images and the natural NRSD set of 165 images.

\textit{2) Implementation Details.}
We achieve our proposed NaDiNet on the PyTorch platform~\cite{PyTorch}, and perform all experiments using an NVIDIA RTX 3090 GPU (24GB memory).
The backbone is initialized with the pre-trained parameters, and the other parts are initialized using the ``Kaiming" method~\cite{InitialWei}.
For all experiments, the input NRSD image and GT are resized to $384\times384$.
We set the batch size, base learning rate, betas, and eps to 12, $1e^{-3}$, (0.9, 0.999), and $1e^{-8}$, respectively.
We train our NaDiNet using the Adam optimizer~\cite{Adam} without data augmentation for 100 epochs until the loss converges.

\begin{figure*}[t!]
    \centering
    \footnotesize
	\begin{overpic}[width=1\textwidth]{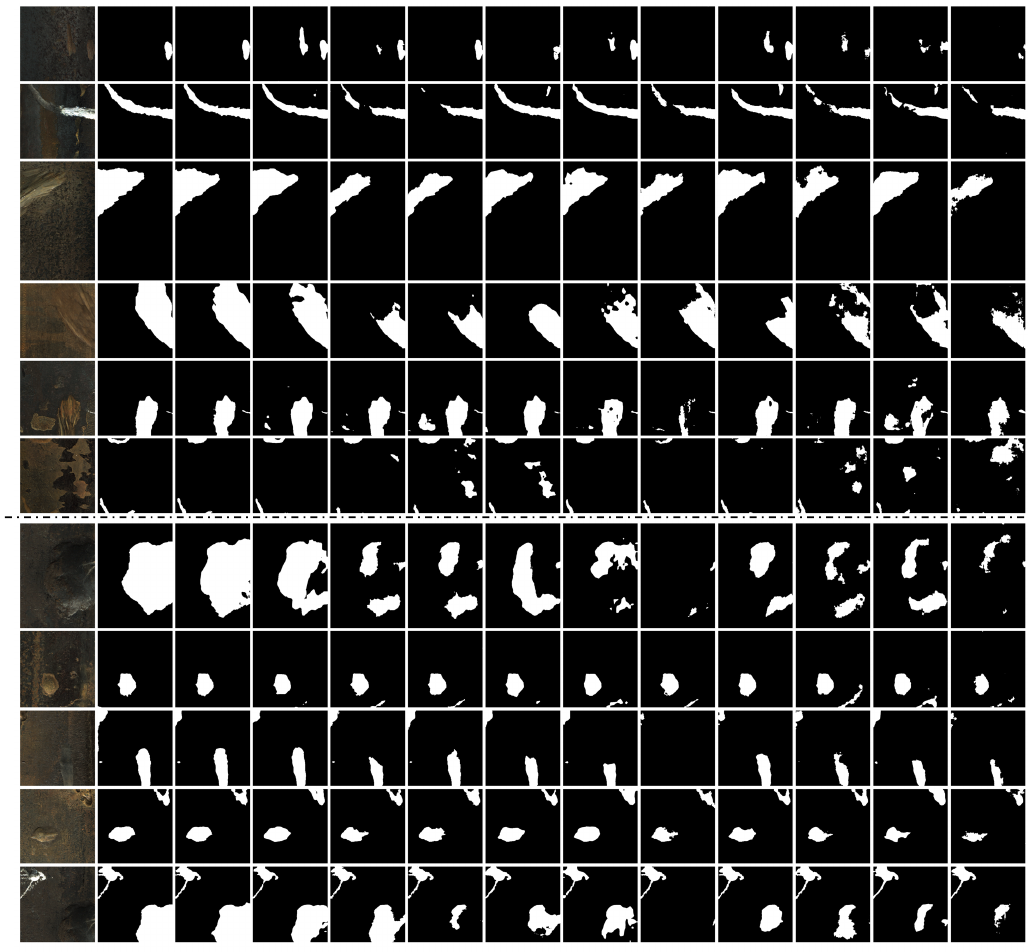}
    \put(-0.03,59.2){ \begin{sideways}{Man-made NRSDs (965)}\end{sideways} }
    \put(-0.03,14.8){ \begin{sideways}{Natural NRSDs (165)}\end{sideways} }

    \put(1.44,-0.5){NRSD image}
    \put(12.20,-0.5){GT}
    \put(18.6,-0.5){Ours-D}
    \put(26.0,-0.5){MCnet}
    \put(33.15,-0.5){EMINet}
    \put(40.65,-0.5){EDRNet}
    \put(47.82,-0.5){TSERNet}
    \put(54.85,-0.5){DeepLabv3}
    \put(63.2,-0.5){DACNet}
    \put(70.42,-0.5){CSEPNet} 
    \put(79.35,-0.5){UNet}
    \put(87.10,-0.5){FCN}
    \put(93.80,-0.5){SegNet}
    
    \end{overpic}
	\caption{Visual comparisons with 10 state-of-the-art methods on two test sets, \ie man-made NRSDs (965) and natural NRSDs (165).
	Please zoom-in for the best view.
    }
    \label{fig:VisualExample}
\end{figure*}
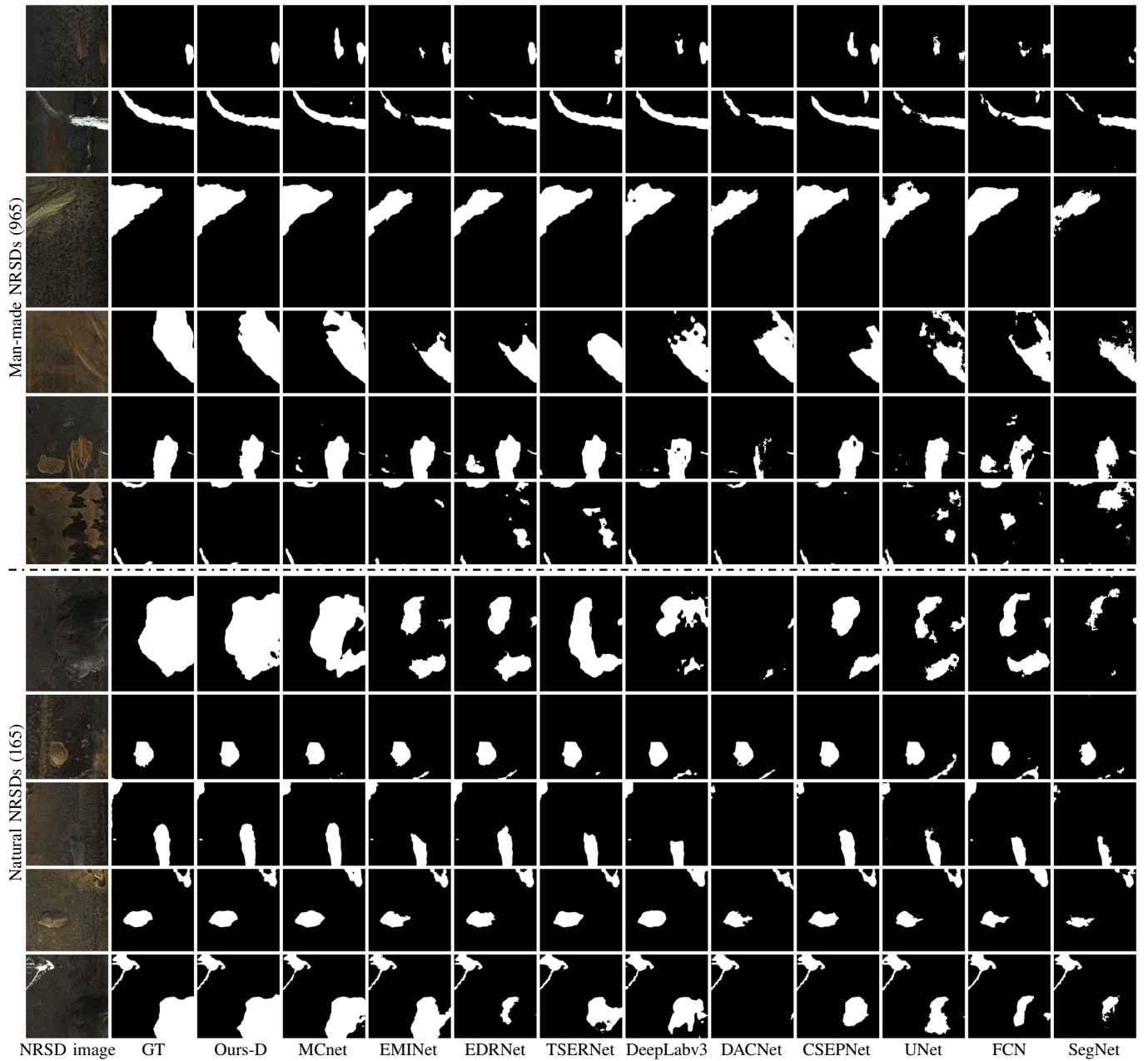

\textit{3) Evaluation Metrics.}
We use two quantitative evaluation metrics to fairly evaluate the segmentation performance of our NaDiNet and all compared methods, including pixel accuracy (PA), intersection over union (IoU), and F1 score.

\textbf{PA} can represent the percentage of correctly classified pixels in the image, and it can be calculated as follows:
\begin{equation}
   \begin{aligned}
	PA = \frac{TP}{TP+FN}, 
    \label{eq:6}
    \end{aligned}
\end{equation}
where $TP$ and $FN$ represent the true positives and false negatives, respectively.

\textbf{IoU} can evaluate the similarities and differences between the predicted segmentation map and the ground truth, and it can be calculated as follows:
\begin{equation}
   \begin{aligned}
	IoU & =   \frac{|\boldsymbol{S}^1 \cap \boldsymbol{G}|}{|\boldsymbol{S}^1 \cup \boldsymbol{G}|},\\
	      & = \frac{TP}{TP+FP+FN},
    \label{eq:6}
    \end{aligned}
\end{equation}
where $FP$ represents the false positives.

\textbf{F1 score} is the harmonic mean of precision and recall, and it can be calculated as follows:
\begin{equation}
   \begin{aligned}
	F1 =   \frac{2 \cdot Precision \cdot Recall}{Precision + Recall}.
    \label{eq:f1}
    \end{aligned}
\end{equation}
The higher these three metrics are, the better the performance.

\subsection{Comparison with State-of-the-arts}
We compare our NaDiNet with a total of 10 state-of-the-art CNN-based methods in three categories.
The first one is the image segmentation method, including SegNet~\cite{2017SegNet}, FCN~\cite{2015FCN}, UNet~\cite{2015Unet}, and Deeplabv3~\cite{DeeplabV3}.
The second one is the saliency detection method for strip steel, including CSEPNet~\cite{2022CSEP}, DACNet\cite{2021DACNet}, TSERNet~\cite{2022TSERNet}, EDRNet\cite{2020EDRNet}, and EMINet~\cite{2021EMINet}.
The last one is the specialized NRSD segmentation method, \ie MCnet~\cite{2021MCNet}.
Except for MCnet dedicated for NRSD segmentation, we retrain the other compared methods with the same dataset and input size (\ie $384\times384$) as our method using their default parameter settings, and test them on the man-made NRSD testing set and the natural NRSD set to get the segmentation maps.
The above compared methods use different backbones for feature extraction, including VGG~\cite{2014VGG16}, ResNet~\cite{2016ResNet}, and DenseNet~\cite{DenseNet}.
Here, for a fair comparison, we also provide three variants of our NaDiNet with backbones of VGG, ResNet, and DenseNet, termed Ours-V, Ours-R, and Ours-D, respectively.

\textit{1) Quantitative Comparison on Man-made and Natural NRSDs.}
We report the quantitative performance of all methods in Tab.~\ref{table:QuantitativeResults}.
Overall, Ours-V, Ours-R, and Ours-D outperform methods with the same backbone as theirs, except for the F1 score of Ours-R on the man-made NRSDs.
Our NaDiNet with different backbones consistently outperforms all compared methods in terms of PA and IoU.
This illustrates the effectiveness of our proposed modules and the flexible adaptability of our methods to different backbones.
Ours-D improves the specialized MCnet by 4.8\% in PA, 3.2\% in IoU, and 2.8\% in F1 score on man-made NRSDs, and by 3.0\% in PA, 2.6\% in IoU, and 2.7\% in F1 score on natural NRSDs.
The performance gain of our method is significant.
Besides, we found the following phenomena.
The performance of the four image segmentation methods (\ie SegNet, FCN, UNet, and Deeplabv3) is downstream of all compared methods, which means that none of the four image segmentation methods are well adapted to the unique scenes of NRSD images.
The performance of saliency detection methods for strip steel is generally better than that of image segmentation methods, which is due to the similarity of strip steel and no-service rail.

\textit{2) Parameter and Speed Comparison.}
Moreover, we measure the network parameters and inference speed (without I/O time) of all methods using the same hardware, and report them in Tab.~\ref{table:QuantitativeResults}.
Overall, under different backbones, our method is among the three methods with the least number of parameters.
This indicates that our method is also efficient, and the effectiveness of our method does not lie in blindly increasing the number of parameters.
In terms of inference speed, Ours-V and Ours-R achieve near real-time inference speeds of 27\textit{fps} and 22\textit{fps}, respectively, which are at the mid-stream level among all comparison methods.
Due to the deep network depth of DenseNet, the inference speeds of Ours-D and MCnet are only 13\textit{fps} and 18\textit{fps}, respectively.

\textit{3) Visual Comparison.}
In Fig.~\ref{fig:VisualExample}, we compare our NaDiNet equipped with the backbone of DenseNet (\ie Ours-D) with all state-of-the-art methods visually on the man-made NRSDs and the natural NRSDs.
There are several challenging scenes in Fig.~\ref{fig:VisualExample}, including multiple defects, tiny defects, the inconsistent defect, and irregular defects.
We can clearly observe that our segmentation maps are close to GTs, while the segmentation maps of other compared methods have the situation of missing defects, incomplete segmentation of defects, or even incorrect segmentation.
The satisfactory segmentation maps of our method are attributed to the enhancement-interaction paradigm we adopted, which first enhances features to make defect regions complete and avoid missing defects, and then performs the cross-level interaction to perceive defect regions of various granularities.
The segmentation maps of image segmentation methods further demonstrate that these methods do not effectively handle NRSD images.
As for the saliency detection methods of strip steel, there is still a certain gap between their segmentation maps and our segmentation maps.

\begin{table}[!t]
\centering
\caption{Quantitative results (\%) of evaluating the contribution of NAM and DIB in NaDiNet.
The subscript is the improved performance compared to Baseline-D.
The best one in each column is \textbf{bold}.
  }
\label{Ablation_module}
\small
\renewcommand{\arraystretch}{1.20}
\renewcommand{\tabcolsep}{1.mm}
\begin{tabular}{c|ccc||cc|cc}
\bottomrule

  \multirow{2}{*}{No.} & \multirow{2}{*}{Baseline-D} & \multirow{2}{*}{NAM} & \multirow{2}{*}{DIB}  
 & \multicolumn{2}{c|}{Man-made (965)}
 & \multicolumn{2}{c}{Natural (165)}   \\
 
    & & & & PA$\uparrow$ & IoU$\uparrow$ & PA$\uparrow$ & IoU$\uparrow$   \\
\hline
\hline
1 &  \Checkmark &                      &                      & 82.0 & 69.3 & 67.3 & 60.4   \\
2 &  \Checkmark & \Checkmark  &                      & 82.6 \tiny{+0.6} & 70.1 \tiny{+0.8} & 68.9 \tiny{+1.6} & 62.2 \tiny{+1.8}    \\
3 &  \Checkmark &                      & \Checkmark  & 83.0 \tiny{+1.0} & 70.5 \tiny{+1.2} & 69.6 \tiny{+2.3} & 62.4 \tiny{+2.0}   \\

\hline
4 &  \Checkmark & \Checkmark  & \Checkmark &  \textbf{84.2 \tiny{+2.2}} & \textbf{71.3 \tiny{+2.0}} & \textbf{72.5 \tiny{+5.2}} & \textbf{65.2 \tiny{+4.8}}  \\
\toprule
\end{tabular}
\end{table}

\subsection{Ablation Studies}
\label{Ablation Studies}
We conduct exhaustive ablation studies on the NRSD-MN dataset to demonstrate the effectiveness of the proposed modules in our NaDiNet. 
Concretely, we analyze these modules from the following three aspects:
1) the contribution of NAM and DIB in NaDiNet,
2) the advantages of NAM over vanilla CAM (\ie NAM \textit{vs.} CAM), and
3) the effectiveness of each part in DIB.
Here, we adopt the DenseNet-201 as the backbone of the feature extractor, and perform all ablation experiments with the same parameter settings as in the Sec.~\ref{sec:ExpProtocol}.

\textit{1) Contribution of NAM and DIB in NaDiNet}.
NAM and DIB are the keys in our NaDiNet.
To evaluate the contribution of NAM and DIB, we design three variants:
1) Baseline-D, in which we remove all NAMs and DIBs and employ a concatenation operator and three convolutional layers to fuse the cross-level features,
2) Baseline-D+NAM, and 
3) Baseline-D+DIB. 
We provide the quantitative results of the above variants in Tab.~\ref{Ablation_module}.

At first glance, we can observe that both modules improve the performance of ``Baseline-D" on two test sets.
Through detailed comparison, NAM boosts the performance of ``Baseline-D" by 0.6\% and 1.6\% in PA and 0.8\% and 1.8\% in IoU, and DIB boosts the performance of ``Baseline-D" by 1.0\% and 2.3\% in PA and 1.2\% and 2.0\% in IoU.
With both modules working together, the full NaDiNet outperforms ``Baseline-D" by a large margin, especially for natural NSRDs, where its advantages reach 5.2\% in PA and 4.8\% in IoU.
This above analysis intuitively shows the strong contribution of NAM and DIB for feature enhancement and feature interaction, and proves the effectiveness of the enhancement-interaction paradigm in our NaDiNet.

\textit{2) Advantages of NAM over vanilla CAM (\ie NAM vs. CAM).}
As presented in Sec.~\ref{sec:NAM}, our NAM is proposed specifically for NSRD images.
By comparing the channel attention maps and features of NAM and CAM in Fig.~\ref{CAM_Visualization} and Fig.~\ref{NAM_Visualization}, we can visually observe the advantages of our NAM.
In this part, we specifically evaluate the advantages of NAM over vanilla CAM from the quantitative perspective. 
To this end, we conduct a variant named \textit{w/ CAM}, which replaces all five NAMs with vanilla CAMs in our NaDiNet.
As the quantitative results shown in the 1$^{st}$ row of Tab.~\ref{table:Ablation_study_1}, we find that our NAM is indeed better for NSRD images than vanilla CAM, which shows that expanding the value of the channel attention map for low-contrast NSRD images can increase the difference of features to achieve effective feature enhancement.
According to the thorough visual and quantitative results, our NAM is superior to vanilla CAM for NSRD images.

\begin{table}[!t]
\centering
\caption{Ablation study  (\%) on evaluating the advantages of NAM and the effectiveness of each part in DIB.
  The best result in each column is \textbf{bold}.
  }
\label{table:Ablation_study_1}
\small
\renewcommand{\arraystretch}{1.20}
\renewcommand{\tabcolsep}{2.9mm}
\begin{tabular}{c|c||cc|cc}
\toprule

 \multirow{2}{*}{No.} 
 & \multirow{2}{*}{Models} 
 & \multicolumn{2}{c|}{Man-made (965)}
 & \multicolumn{2}{c}{Natural (165)}   \\
 
     & & PA$\uparrow$ & IoU$\uparrow$ & PA$\uparrow$ & IoU$\uparrow$   \\
\midrule
1 & \textit{w/ CAM} 			& 83.6 & 70.6 & 71.4 & 64.0\\ 

\hline
2 & \textit{w/o SI}& 83.4 & 70.9 & 70.0 & 63.4\\ 
3 & \textit{w/ large-scale}	& 83.6 & 70.7 & 70.9 & 63.3\\ 
4 & \textit{w/ small-scale}	& 83.7 & 70.4 & 70.4 & 62.2\\ 
5 & \textit{w/o PE}	& 83.5 & 70.6 & 69.7 & 62.4\\ 

\hline
6 & \textbf{Ours-D} 			&  \textbf{84.2} & \textbf{71.3} & \textbf{72.5} & \textbf{65.2}  \\
\toprule
\end{tabular}
\end{table}

\textit{3) Effectiveness of Each Part in DIB.}
DIB is in charge of the interaction of the cross-level features, and contains several parts, such as the large-scale and small-scale interaction branches.
In the following, we analyze their effectiveness one by one.

First, to prove the necessity of the scale interaction in both branches, we forego concatenating $\boldsymbol{f}^{i}_\mathrm{na}$ and $\boldsymbol{f}^{i+1}_\mathrm{di} $/$\boldsymbol{f}_\mathrm{gc}$, and perform the progressive ensemble on $\boldsymbol{f}^{i}_\mathrm{na}$ and $\boldsymbol{f}^{i+1}_\mathrm{di} $/$\boldsymbol{f}_\mathrm{gc}$ individually rather than on their combination in \textit{w/ SI}.  
As results shown in the 2$^{nd}$ row of Tab.~\ref{table:Ablation_study_1}, the scale interaction is integral for cross-level fusion.
Then, to prove the effectiveness of performing interactions at dual scales, we provide two variants.
The first one \textit{w/ large-scale} only retains the large-scale interaction branch (\ie the left one in Fig.~\ref{DIB_structure}), while the second one \textit{w/ small-scale} only retains the small-scale interaction branch (\ie the right one in Fig.~\ref{DIB_structure}).
The quantitative performance reported in Tab.~\ref{table:Ablation_study_1} shows that two different-scale interaction branches are indispensable, indicating that interaction at dual scales can indeed capture features of different granularities that are beneficial to NSRD segmentation.
Last, we provide a variant \textit{w/ PE} to evaluate the superiority of the progressive ensemble manner in both branches, that is, we directly modulate $\boldsymbol{{f}}^{i}_\mathrm{b}$ and $\boldsymbol{{f}}^{i}_\mathrm{s}$ with one CBAM.
According to the results of the 5$^{th}$ row in Tab.~\ref{table:Ablation_study_1}, we conclude that the progressive ensemble manner is good for NSRD segmentation, and it can capture multi-scale information to effectively characterize defect regions of different sizes.
The above analysis proves that each part of our DIB is effective.

\section{Conclusion}
\label{sec:con}
In this paper, we propose a novel specialized NaDiNet for NSRD segmentation, following the enhancement-interaction paradigm.
To address the unique low-contrast textures of NSRD images, we update vanilla CAM to NAM for intra-level feature enhancement, which directly calculates the normalized dependencies between channels to expand the value of attention maps.
Furthermore, we achieve the inter-level feature interaction in DIB to capture the fine-grained and coarse-grained information of defects, thereby handling scenes with defects of various shapes and scales.
In order to accelerate loss convergence and improve the representation ability of features, we adopt the deep supervision strategy during the network training phase.
Comprehensive quantitative and visual comparisons, as well as ablation experiments, prove the superior performance of our NaDiNet and the flexibility and advantages of our NAM and DIB.
In our future work, we plan to take the segmentation map generated by our NaDiNet as a prompt of the popular Segment Anything Model (SAM)\cite{2023SAM} and utilize the powerful segmentation ability of SAM to generate a more accurate segmentation map.



\ifCLASSOPTIONcaptionsoff
  \newpage
\fi

\bibliographystyle{IEEEtran}
\bibliography{myRef}

%



%

\end{document}